\documentclass{article}

\usepackage{graphicx}
\usepackage{amsmath}
\usepackage{cite}
\usepackage{booktabs}
\usepackage{hyperref}
\usepackage{xcolor}
\usepackage[accepted]{./icml2025}
\usepackage{xurl}

\icmltitlerunning{Comparative Evaluation of Prompting and Fine-Tuning for Applying LLMs to Grid-Structured Geospatial Data}

\begin{document}

\twocolumn[

\icmltitle{Comparative Evaluation of Prompting and Fine-Tuning for Applying Large Language Models to Grid-Structured Geospatial Data}

\begin{icmlauthorlist}
\icmlauthor{Akash Dhruv}{1}
\icmlauthor{Yangxinyu Xie}{1,2}
\icmlauthor{Jordan Branham}{3}
\icmlauthor{Tanwi Mallick}{1}
\end{icmlauthorlist}

\icmlaffiliation{1}{Mathematics and Computer Science, Argonne National Laboratory, Lemont, IL, USA}
\icmlaffiliation{2}{The Wharton School, University of Pennsylvania, Philadelphia, PA, USA}
\icmlaffiliation{3}{Decision and Infrastructure, Argonne National Laboratory, Lemont, IL, USA}

\icmlcorrespondingauthor{Akash Dhruv}{}

\icmlkeywords{LLMs, Generative AI, Grid-Structured Datasets, Atmospheric Data, Geospatial Reasoning, Weather Resilience}

\vskip 0.3in
]


\printAffiliationsAndNotice{}  

\begin{abstract}
This paper presents a comparative study of large language models (LLMs) in interpreting grid-structured geospatial data. We evaluate the performance of a base model through structured prompting and contrast it with a fine-tuned variant trained on a dataset of user-assistant interactions. Our results highlight the strengths and limitations of zero-shot prompting and demonstrate the benefits of fine-tuning for structured geospatial and temporal reasoning.
\end{abstract}

\section{Introduction}
Geospatial datasets are essential for weather forecasting and resilience planning. They provide key variables such as temperature, precipitation, wind, and humidity, organized over structured grids indexed by latitude and longitude. These datasets support a wide range of applications, from infrastructure design to emergency preparedness.

The data originate from diverse sources, including ground-based observations, satellite retrievals, reanalysis products, and numerical models. Reanalysis datasets such as ERA5 \cite{https://doi.org/10.1002/qj.3803} and MERRA-2 \cite{TheModernEraRetrospectiveAnalysisforResearchandApplicationsVersion2MERRA2} combine observations with physical models to produce coherent, gap-filled time series. Others, like CHIRPS \cite{Funk2015} and APHRODITE \cite{Yatagai2012}, use statistical methods to estimate values in regions where direct measurements are sparse. Satellite products offer broad spatial coverage but often require careful calibration, and high-resolution atmospheric models, such as WRF\footnote{https://github.com/wrf-model/WRF}, simulate physical processes to generate detailed, gridded outputs for both forecasting and weather projections.

Although structurally regular, these datasets present significant challenges for integration into modern AI workflows. They are typically represented as multidimensional arrays or tables indexed by space and time, 
making them difficult for standard foundation models to parse, interpret, and reason over due to their dense numerical content and complex spatiotemporal dependencies.
More importantly, they differ substantially from the natural language formats that large language models (LLMs) are designed to process. Training LLMs directly on large volumes of numerical weather data is impractical due to memory limitations, restricted context windows, and inefficiencies that arise from learning without structure-aware encoding.

To date, most LLM applications in science have focused on text-centric domains such as biomedicine and code generation \cite{dhruv2025leveraginglargelanguagemodels}. Spatially organized scientific data, including atmospheric datasets, have received less attention. These datasets rely on spatial relationships, physical units, and exact numerical values, which are difficult to represent in plain text or linear sequences. Nonetheless, recent research is beginning to address this gap. Google's work on geospatial foundation models \cite{google2024geospatial} and hybrid approaches that combine LLMs with spatial encoders \cite{symufolk2024climate} point to promising directions. However, effectively preserving the spatial structure and physical integrity of atmospheric data within LLM frameworks remains an open problem.

This challenge becomes evident when integrating datasets like ClimRR\footnote{https://climrr.anl.gov}, which are designed to support urban planning and hazard adaptation across the United States, into agentic LLM workflows for deployment in GPT-like chat interfaces. ClimRR provides projections of extreme temperature, precipitation, wind speed, humidity, and fire weather indices under two Representative Concentration Pathways (RCP 4.5 and 8.5). Generated using WRF, the dataset presents a tabular representation of atmospheric variables indexed by unique crossmodel tags corresponding to spatial locations across the U.S. This makes ClimRR an ideal test case for evaluating how effectively LLMs can interpret real-world gridded atmospheric data.

{In agentic workflows, where multiple specialized models operate in coordination, small models with fewer parameters (8B or less) are particularly advantageous. They provide a favorable balance between performance and resource efficiency, enabling deployment across many tasks without exceeding memory or computational limits}. Relying on ever-larger models to ingest raw data would compromise one of the key benefits of LLMs in these workflows: their ability to remain lightweight, modular, and adaptable.

In this study, we investigate how LLMs can be adapted to reason over structured datasets like ClimRR. We explore both prompt-based and fine-tuned approaches, targeting tasks such as value extraction, scenario comparison, and domain-informed interpretation. Building on an earlier work with WildfireGPT \cite{xie2025wildfiregpttailoredlargelanguage, xie2025rag}, which focused on spatial reasoning for wildfire risk, we expand our scope to a wider set of atmospheric variables and to the broader challenge of scientific reasoning grounded in spatial context. {Our approach emphasizes 
small, open-source models with low parameter count, such as LLaMA 3.1 8B  \cite{grattafiori2024llama}, to support fast inference, low-latency responses, and scalable deployment. These lightweight models enable fast responses for time-sensitive tasks like early warnings and situational assessments, while supporting efficient deployment and integration into domain-specific decision-support tools.}

\section{Problem Setting and Dataset}

Our central research question is whether large language models can meaningfully interpret and reason over spatially and temporally structured geospatial data, as outlined in the previous section. Unlike typical LLM tasks focused primarily on natural language, this challenge requires understanding complex spatial relationships and temporal patterns that characterize atmospheric datasets and the physical processes they represent.

To systematically investigate LLM capabilities on such data, we define a set of core tasks for the models:
(1) Answering user queries about specific grid cells using provided input context;
(2) Distinguishing between mid-century and end-century timeframes;
(3) Summarizing weather trends relative to user-specified regions such as states or counties;
(4) Providing information for one or both emissions scenarios (RCP 4.5 and RCP 8.5), depending on the query.

To support these tasks, we developed a curated dataset of structured JSON records, each simulating a user query paired with relevant data and a reference answer. Each record includes:
(1) A unique grid cell identifier (e.g., R073C493) corresponding to a location within the continental U.S.;
(2) Atmospheric variables spanning three time periods: historical (1971–2000), mid-century (2041–2070), and end-century (2071–2100);
(3) Projections under both RCP 4.5 and RCP 8.5 emissions scenarios, along with regional aggregates;
(4) A ground-truth response combining precise value extraction with natural language interpretation.
\begin{figure*}[ht]
\centering
\includegraphics[width=1.0\linewidth]{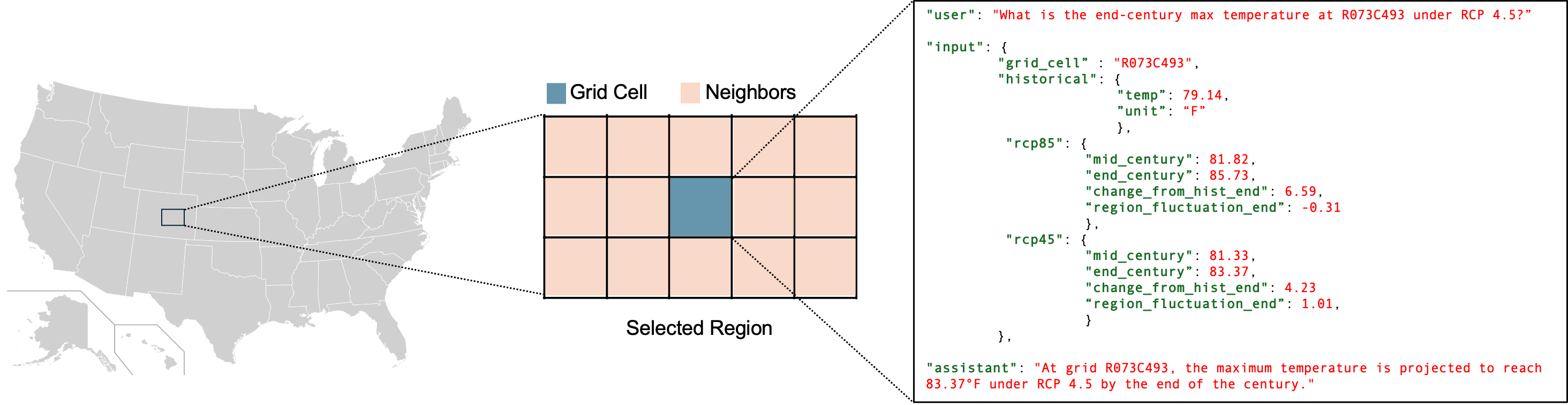}
\vspace{-0.7cm}
\caption{Schematic showing the mapping of gridded ClimRR data over United States. Each grid cell is assigned an alphanumeric tag (e.g., R073C493) and contains atmospheric variable values in tabular form. These values can be transformed into a user–input–assistant format suitable for prompting and fine-tuning the language model.}
\label{fig:json_example}
\end{figure*}
\begin{figure*}[ht]
\centering
\vspace{0.25cm}
\includegraphics[width=0.65\linewidth]{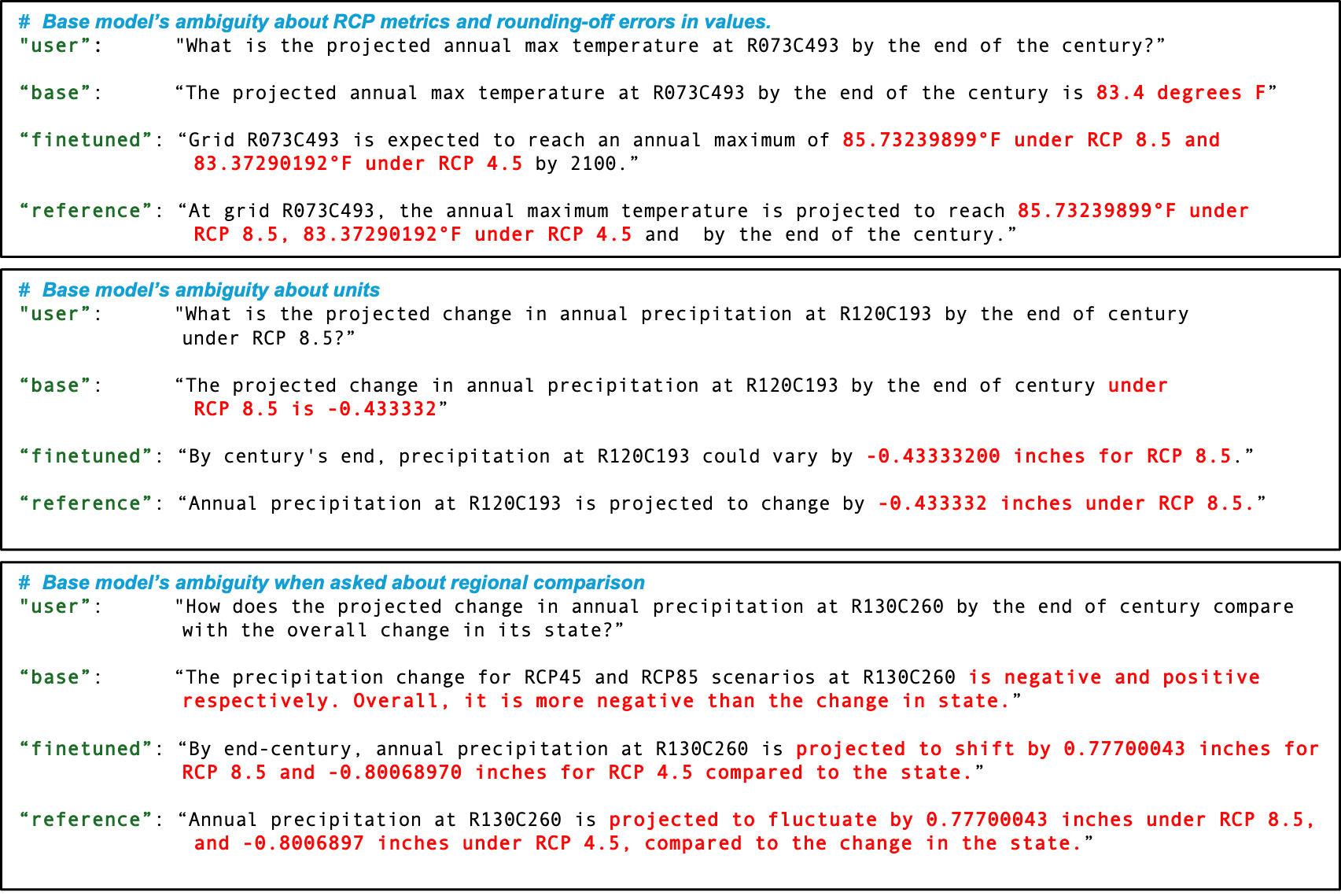}
\vspace{-0.4cm}
\caption{
Comparison of reference outputs with responses from the base and fine-tuned models to user queries, highlighting differences (shown in red) in accuracy and reasoning. These examples illustrate common ambiguities in base model responses to geospatial climate queries, including challenges in referencing correct RCP scenarios, handling measurement units, and making accurate regional comparisons. In contrast, fine-tuned models show improved alignment with reference answers across all categories, demonstrating an enhanced understanding of domain-specific nuances.}
\label{fig:output_comparison}
\end{figure*}
This dataset was generated via a semi-automated pipeline that queries the original ClimRR data source, extracts relevant variables for selected grid cells and time periods, and produces question-answer pairs based on predefined templates. We then used GPT-4 to introduce linguistic variation by rephrasing queries and responses, helping to remove template-induced patterns and increase diversity. This approach enables the creation of a large, consistent dataset that preserves the complex spatial and temporal structure of the original data. Currently, the dataset comprises approximately 120 examples covering a range of queries from simple value retrieval to scenario comparison. Figure \ref{fig:json_example} illustrates how this dataset supports interaction on the ClimRR web interface, where a user selects a region and queries an LLM about specific grid cells. These interactions rely on the structured JSON format described above and form the foundation for evaluating model performance across several core tasks:

\begin{itemize}
    \item \textbf{Variable Retrieval}: Can the model correctly extract the value of a specified variable for a given grid cell and time period?
    \item \textbf{Trend Analysis}: Can it summarize how a variable changes over time within a spatial context?
    \item \textbf{Scenario Comparison}: Can it identify and describe differences between emissions scenarios (RCP 4.5 vs. RCP 8.5)?
    \item \textbf{Contextual Interpretation}: Can it understand a user query and provide context about minima or maxima of projections relative to the relevant region?
\end{itemize}

It is important to note that the extraction of precise grid cell values and preparation of input context are handled by a separate supervising system outside the LLM under evaluation. Our focus is to test whether, given a user query and relevant input, the LLM can accurately extract correct values with units and deliver appropriate, domain-aligned answers.

Standard LLM tokenization and training approaches struggle with structured scientific data because numbers are treated as opaque tokens and spatial relationships are flattened. Our dataset bridges this gap, enabling rigorous evaluation of LLMs’ ability to handle structured, spatially indexed, and numerically precise data.

Overall, this framework establishes a benchmark to assess LLM performance on relevant tasks that require spatial awareness, numerical accuracy, and scientific interpretability, capabilities essential for real-world applications in weather resilience and adaptation planning.

\section{Prompting Methodology}
We began our evaluation by testing how well a base LLM (without any task-specific fine-tuning) could reason over structured inputs when guided by carefully designed prompts. This approach reflects a realistic use case where users interact with off-the-shelf LLMs through natural language queries supplemented by structured data. Each prompt consisted of two components:
\begin{itemize}
    \item A structured \textbf{input} block containing grid-indexed data for a single cell, including historical values, projections under RCP 4.5 and RCP 8.5 scenarios, and fluctuations relative to the region (state, county, etc.) aggregate.
    \item A natural language \textbf{user} query requesting specific information, such as projected values, temporal trends, or scenario comparisons.
\end{itemize}
We constructed a suite of prompt templates and evaluated model performance on a test subset comprising 10\% of the dataset. Even without fine-tuning, the base model demonstrated promising capabilities. It reliably extracted values from JSON-style inputs and answered straightforward questions about specific locations and time periods. In many cases, it successfully compared emissions scenarios and identified temporal trends. However, several limitations were observed:
\begin{itemize}
    \item The model occasionally introduced rounding or interpolation errors, resulting in minor discrepancies compared to source data.
    \item It sometimes ignored units of the physical quantities, and provided ambiguous response when asked to provide comparative data related to overall region (state, county, etc.). 
    \item When the user query did not specify an emissions scenario, the model often defaulted to a single, arbitrary scenario without including both RCP 4.5 and 8.5 or clearly indicating which one it was referencing.
\end{itemize}
These findings suggest that while structured prompting enables basic data retrieval and comparison, it falls short in delivering consistent contextual fluency and robust scientific synthesis. Nonetheless, the base model’s performance on cleanly structured prompts establishes a valuable baseline and indicates that further gains are achievable through fine-tuning and improved data representation strategies.
%

\section{Fine-Tuning and Experimental Design}

To enhance model performance, we fine-tuned a 8B parameter open-weight {LLaMA 3.1 \cite{grattafiori2024llama,touvron2023llamaopenefficientfoundation}} language model using approximately 100 user–input–assistant examples. Each example was designed to teach the model to: (1) Interpret grid-specific weather projections; (2) Compare historical and future scenarios (e.g., RCP4.5 vs. RCP8.5); and (3) Generate contextually accurate and numerically grounded responses.

Fine-tuning used parameter-efficient Low-Rank Adaptation (LoRA) \cite{hu2021loralowrankadaptationlarge} via the Unsloth-AI API \cite{unsloth}, with rank 8 and scaling factor 16 to update a small subset of parameters. 8-bit quantization with BitsAndBytes \cite{dettmers20228bitoptimizersblockwisequantization} reduced memory usage, while mixed precision training with {bfloat16} improved efficiency and stability \cite{micikevicius2018mixedprecisiontraining}. Inputs were fixed at 2048 tokens \cite{vaswani2023attentionneed}, and training used small batches with gradient accumulation (effective batch size of 8) \cite{ott2018scalingneuralmachinetranslation}. Optimization employed AdamW \cite{loshchilov2019decoupledweightdecayregularization} with cosine decay and warm-up \cite{loshchilov2017sgdrstochasticgradientdescent}. Training was monitored using Weights \& Biases (wandb)\footnote{https://wandb.ai} for reproducibility. The resulting fine-tuned model exhibited improved spatial reasoning and robustness in producing value-aware responses grounded in the underlying data. The GitHub repository containing our training scripts and datasets is publicly available\footnote{{https://github.com/Lab-Notebooks/ARAIA-Model-Finetuning}}.

\section{Comparative Evaluation}
\begin{table}[h]
\centering
\begin{tabular}{lcc}
\hline
\textbf{Model} & \textbf{Similarity Score} & \textbf{Accuracy Score} \\
\hline
Base Model & 0.8335 & 0.2889 \\
Finetuned Model & 0.8954 & 1.0 \\
\hline
\end{tabular}
\vspace{-0.25cm}
\caption{ Summarizes model performance, showing that the base model achieves moderate semantic similarity but low accuracy due to frequent errors in values, units, and scenario interpretation. In contrast, the fine-tuned model achieves higher similarity and accuracy, indicating precise and consistent outputs.}
\label{tab:model_comparison}
\end{table}
Table \ref{tab:model_comparison} compares the base and fine-tuned LLMs using cosine similarity (semantic alignment) and accuracy (exact value correctness). We calculated semantic similarity using the all-MiniLM-L6-v2 model from SentenceTransformer library, comparing sentence embeddings of model outputs and reference responses via cosine similarity. For accuracy, {we wrote a Python script that used regex to extract key elements grid cell, variable, units, RCP scenario, and values from each response}. We assigned scores based on exact (1.0), partial (0.5), or no match (0.0), and computed an average accuracy score with equal weighting across components. The base model achieved a similarity of 0.8335 and low accuracy (0.2889), indicating general understanding but frequent numeric or scenario-related errors. These included inconsistent emissions handling, unit confusion, and rounding.

Fine-tuning significantly improved both similarity and score to 0.8954 and 1.0 respectively, showing better extraction of exact values, scenario awareness, and unit consistency. While structured prompts alone enabled basic reasoning, fine-tuning was essential for high-precision, domain-aligned answers. A comparison of select results in shown in Figure \ref{fig:output_comparison}.

\section{Conclusion and Future Work}
Our study demonstrates that LLMs can effectively reason over structured data when guided by well-designed prompts and fine-tuned on task-relevant examples. The base model showed initial promise but lacked consistency in handling emissions scenarios, units, and precise values. Fine-tuning substantially improved performance across all key metrics.

Looking ahead, we plan to broaden the dataset to include more complex and nuanced queries. We are also developing a real-time agentic workflow where an external system extracts relevant data from the ClimRR API, and the LLM interprets it on demand. This will support interactive, location-aware weather exploration for applications in planning, resilience, and public engagement.

\section{Acknowledgments}
The submitted manuscript has been created by UChicago Argonne, LLC, Operator of Argonne National Laboratory (“Argonne”). Argonne's work was supported by the U.S. Department of Energy, Grid Deployment Office, under contract DE-AC02-06CH11357.

The U.S. Government retains for itself, and others acting on its behalf, a paid-up nonexclusive, irrevocable worldwide license in said article to reproduce, prepare derivative works, distribute copies to the public, and perform publicly and display publicly, by or on behalf of the Government. The Department of Energy will provide public access to these results of federally sponsored research in accordance with the DOE Public Access Plan. http://energy.gov/downloads/doe-public-access-plan.

\bibliographystyle{icml2025}
\bibliography{references}

\begin{thebibliography}{19}
\providecommand{\natexlab}[1]{#1}
\providecommand{\url}[1]{\texttt{#1}}
\expandafter\ifx\csname urlstyle\endcsname\relax
  \providecommand{\doi}[1]{doi: #1}\else
  \providecommand{\doi}{doi: \begingroup \urlstyle{rm}\Url}\fi

\bibitem[Daniel~Han \& team(2023)Daniel~Han and team]{unsloth}
Daniel~Han, M.~H. and team, U.
\newblock Unsloth, 2023.
\newblock URL \url{http://github.com/unslothai/unsloth}.

\bibitem[Dettmers et~al.(2022)Dettmers, Lewis, Shleifer, and
  Zettlemoyer]{dettmers20228bitoptimizersblockwisequantization}
Dettmers, T., Lewis, M., Shleifer, S., and Zettlemoyer, L.
\newblock 8-bit optimizers via block-wise quantization, 2022.
\newblock URL \url{https://arxiv.org/abs/2110.02861}.

\bibitem[Dhruv \& Dubey(2025)Dhruv and
  Dubey]{dhruv2025leveraginglargelanguagemodels}
Dhruv, A. and Dubey, A.
\newblock Leveraging large language models for code translation and software
  development in scientific computing, 2025.
\newblock URL \url{https://arxiv.org/abs/2410.24119}.

\bibitem[Funk et~al.(2015)Funk, Peterson, Landsfeld, Pedreros, Verdin, Shukla,
  Husak, Rowland, Harrison, Hoell, and Michaelsen]{Funk2015}
Funk, C., Peterson, P., Landsfeld, M., Pedreros, D., Verdin, J., Shukla, S.,
  Husak, G., Rowland, J., Harrison, L., Hoell, A., and Michaelsen, J.
\newblock The climate hazards infrared precipitation with stations—a new
  environmental record for monitoring extremes.
\newblock \emph{Scientific Data}, 2\penalty0 (1):\penalty0 150066, 2015.
\newblock ISSN 2052-4463.
\newblock \doi{10.1038/sdata.2015.66}.
\newblock URL \url{https://doi.org/10.1038/sdata.2015.66}.

\bibitem[Gelaro et~al.(2017)Gelaro, McCarty, Suárez, Todling, Molod, Takacs,
  Randles, Darmenov, Bosilovich, Reichle, Wargan, Coy, Cullather, Draper,
  Akella, Buchard, Conaty, da~Silva, Gu, Kim, Koster, Lucchesi, Merkova,
  Nielsen, Partyka, Pawson, Putman, Rienecker, Schubert, Sienkiewicz, and
  Zhao]{TheModernEraRetrospectiveAnalysisforResearchandApplicationsVersion2MERRA2}
Gelaro, R., McCarty, W., Suárez, M.~J., Todling, R., Molod, A., Takacs, L.,
  Randles, C.~A., Darmenov, A., Bosilovich, M.~G., Reichle, R., Wargan, K.,
  Coy, L., Cullather, R., Draper, C., Akella, S., Buchard, V., Conaty, A.,
  da~Silva, A.~M., Gu, W., Kim, G.-K., Koster, R., Lucchesi, R., Merkova, D.,
  Nielsen, J.~E., Partyka, G., Pawson, S., Putman, W., Rienecker, M., Schubert,
  S.~D., Sienkiewicz, M., and Zhao, B.
\newblock The modern-era retrospective analysis for research and applications,
  version 2 (merra-2).
\newblock \emph{Journal of Climate}, 30\penalty0 (14):\penalty0 5419 -- 5454,
  2017.
\newblock \doi{10.1175/JCLI-D-16-0758.1}.
\newblock URL
  \url{https://journals.ametsoc.org/view/journals/clim/30/14/jcli-d-16-0758.1.xml}.

\bibitem[{Google Research}(2025)]{google2024geospatial}
{Google Research}.
\newblock Geospatial reasoning: Unlocking insights with generative ai and
  multiple foundation models.
\newblock
  \url{https://research.google/blog/geospatial-reasoning-unlocking-insights-with-generative-ai-and-multiple-foundation-models/},
  2025.

\bibitem[Grattafiori et~al.(2024)Grattafiori, Dubey, Jauhri, Pandey, Kadian,
  Al-Dahle, Letman, Mathur, Schelten, Vaughan, et~al.]{grattafiori2024llama}
Grattafiori, A., Dubey, A., Jauhri, A., Pandey, A., Kadian, A., Al-Dahle, A.,
  Letman, A., Mathur, A., Schelten, A., Vaughan, A., et~al.
\newblock The llama 3 herd of models.
\newblock \emph{arXiv preprint arXiv:2407.21783}, 2024.

\bibitem[Hersbach et~al.(2020)Hersbach, Bell, Berrisford, Hirahara, Horányi,
  Muñoz-Sabater, Nicolas, Peubey, Radu, Schepers, Simmons, Soci, Abdalla,
  Abellan, Balsamo, Bechtold, Biavati, Bidlot, Bonavita, De~Chiara, Dahlgren,
  Dee, Diamantakis, Dragani, Flemming, Forbes, Fuentes, Geer, Haimberger,
  Healy, Hogan, Hólm, Janisková, Keeley, Laloyaux, Lopez, Lupu, Radnoti,
  de~Rosnay, Rozum, Vamborg, Villaume, and
  Thépaut]{https://doi.org/10.1002/qj.3803}
Hersbach, H., Bell, B., Berrisford, P., Hirahara, S., Horányi, A.,
  Muñoz-Sabater, J., Nicolas, J., Peubey, C., Radu, R., Schepers, D., Simmons,
  A., Soci, C., Abdalla, S., Abellan, X., Balsamo, G., Bechtold, P., Biavati,
  G., Bidlot, J., Bonavita, M., De~Chiara, G., Dahlgren, P., Dee, D.,
  Diamantakis, M., Dragani, R., Flemming, J., Forbes, R., Fuentes, M., Geer,
  A., Haimberger, L., Healy, S., Hogan, R.~J., Hólm, E., Janisková, M.,
  Keeley, S., Laloyaux, P., Lopez, P., Lupu, C., Radnoti, G., de~Rosnay, P.,
  Rozum, I., Vamborg, F., Villaume, S., and Thépaut, J.-N.
\newblock The era5 global reanalysis.
\newblock \emph{Quarterly Journal of the Royal Meteorological Society},
  146\penalty0 (730):\penalty0 1999--2049, 2020.
\newblock \doi{https://doi.org/10.1002/qj.3803}.
\newblock URL
  \url{https://rmets.onlinelibrary.wiley.com/doi/abs/10.1002/qj.3803}.

\bibitem[Hu et~al.(2021)Hu, Shen, Wallis, Allen-Zhu, Li, Wang, Wang, and
  Chen]{hu2021loralowrankadaptationlarge}
Hu, E.~J., Shen, Y., Wallis, P., Allen-Zhu, Z., Li, Y., Wang, S., Wang, L., and
  Chen, W.
\newblock Lora: Low-rank adaptation of large language models, 2021.
\newblock URL \url{https://arxiv.org/abs/2106.09685}.

\bibitem[Loshchilov \& Hutter(2017)Loshchilov and
  Hutter]{loshchilov2017sgdrstochasticgradientdescent}
Loshchilov, I. and Hutter, F.
\newblock Sgdr: Stochastic gradient descent with warm restarts, 2017.
\newblock URL \url{https://arxiv.org/abs/1608.03983}.

\bibitem[Loshchilov \& Hutter(2019)Loshchilov and
  Hutter]{loshchilov2019decoupledweightdecayregularization}
Loshchilov, I. and Hutter, F.
\newblock Decoupled weight decay regularization, 2019.
\newblock URL \url{https://arxiv.org/abs/1711.05101}.

\bibitem[Micikevicius et~al.(2018)Micikevicius, Narang, Alben, Diamos, Elsen,
  Garcia, Ginsburg, Houston, Kuchaiev, Venkatesh, and
  Wu]{micikevicius2018mixedprecisiontraining}
Micikevicius, P., Narang, S., Alben, J., Diamos, G., Elsen, E., Garcia, D.,
  Ginsburg, B., Houston, M., Kuchaiev, O., Venkatesh, G., and Wu, H.
\newblock Mixed precision training, 2018.
\newblock URL \url{https://arxiv.org/abs/1710.03740}.

\bibitem[Ott et~al.(2018)Ott, Edunov, Grangier, and
  Auli]{ott2018scalingneuralmachinetranslation}
Ott, M., Edunov, S., Grangier, D., and Auli, M.
\newblock Scaling neural machine translation, 2018.
\newblock URL \url{https://arxiv.org/abs/1806.00187}.

\bibitem[Symufolk(2025)]{symufolk2024climate}
Symufolk.
\newblock Llms for climate data analytics.
\newblock \url{https://symufolk.com/llm-for-climate-data-analytics/}, 2025.

\bibitem[Touvron et~al.(2023)Touvron, Lavril, Izacard, Martinet, Lachaux,
  Lacroix, Rozière, Goyal, Hambro, Azhar, Rodriguez, Joulin, Grave, and
  Lample]{touvron2023llamaopenefficientfoundation}
Touvron, H., Lavril, T., Izacard, G., Martinet, X., Lachaux, M.-A., Lacroix,
  T., Rozière, B., Goyal, N., Hambro, E., Azhar, F., Rodriguez, A., Joulin,
  A., Grave, E., and Lample, G.
\newblock Llama: Open and efficient foundation language models, 2023.
\newblock URL \url{https://arxiv.org/abs/2302.13971}.

\bibitem[Vaswani et~al.(2023)Vaswani, Shazeer, Parmar, Uszkoreit, Jones, Gomez,
  Kaiser, and Polosukhin]{vaswani2023attentionneed}
Vaswani, A., Shazeer, N., Parmar, N., Uszkoreit, J., Jones, L., Gomez, A.~N.,
  Kaiser, L., and Polosukhin, I.
\newblock Attention is all you need, 2023.
\newblock URL \url{https://arxiv.org/abs/1706.03762}.

\bibitem[Xie et~al.(2025{\natexlab{a}})Xie, Jiang, Mallick, Bergerson,
  Hutchison, Verner, Branham, Alexander, Ross, Feng, Levy, Su, and
  Taylor]{xie2025wildfiregpttailoredlargelanguage}
Xie, Y., Jiang, B., Mallick, T., Bergerson, J.~D., Hutchison, J.~K., Verner,
  D.~R., Branham, J., Alexander, M.~R., Ross, R.~B., Feng, Y., Levy, L.-A., Su,
  W., and Taylor, C.~J.
\newblock Wildfiregpt: Tailored large language model for wildfire analysis,
  2025{\natexlab{a}}.
\newblock URL \url{https://arxiv.org/abs/2402.07877}.

\bibitem[Xie et~al.(2025{\natexlab{b}})Xie, Jiang, Mallick, Bergerson,
  Hutchison, Verner, Branham, Alexander, Ross, Feng, et~al.]{xie2025rag}
Xie, Y., Jiang, B., Mallick, T., Bergerson, J.~D., Hutchison, J.~K., Verner,
  D.~R., Branham, J., Alexander, M.~R., Ross, R.~B., Feng, Y., et~al.
\newblock A rag-based multi-agent llm system for natural hazard resilience and
  adaptation.
\newblock \emph{arXiv preprint arXiv:2504.17200}, 2025{\natexlab{b}}.

\bibitem[Yatagai et~al.(2012)Yatagai, Kamiguchi, Arakawa, Hamada, Yasutomi, and
  Kitoh]{Yatagai2012}
Yatagai, A., Kamiguchi, K., Arakawa, O., Hamada, A., Yasutomi, N., and Kitoh,
  A.
\newblock Aphrodite: Constructing a long-term daily gridded precipitation
  dataset for asia based on a dense network of rain gauges.
\newblock \emph{Bulletin of the American Meteorological Society}, 93\penalty0
  (9):\penalty0 1401--1415, 2012.
\newblock \doi{10.1175/BAMS-D-11-00122.1}.
\newblock URL \url{https://doi.org/10.1175/BAMS-D-11-00122.1}.

\end{thebibliography}

\end{document}